\documentclass[12pt,twoside]{article}
\usepackage{amsmath,amssymb,amsthm}
\edef\,{\thinspace} \edef\;{\thickspace} \edef\!{\negthinspace} 
\def\dispmuskip{\thinmuskip= 3mu plus 0mu minus 2mu \medmuskip=  4mu plus 2mu minus 2mu \thickmuskip=5mu plus 5mu minus 2mu}
\def\textmuskip{\thinmuskip= 0mu                    \medmuskip=  1mu plus 1mu minus 1mu \thickmuskip=2mu plus 3mu minus 1mu}
\textmuskip
\def\beq{\dispmuskip\begin{equation}}    \def\eeq{\end{equation}\textmuskip}
\def\beqn{\dispmuskip\begin{displaymath}}\def\eeqn{\end{displaymath}\textmuskip}
\def\bqa{\dispmuskip\begin{eqnarray}}    \def\eqa{\end{eqnarray}\textmuskip}
\def\bqan{\dispmuskip\begin{eqnarray*}}  \def\eqan{\end{eqnarray*}\textmuskip}
\def\paradot#1{\vspace{1.3ex plus 0.5ex minus 0.5ex}\noindent{\bf{#1.}}}

\newtheorem{theorem}{Theorem}
\newtheorem{corollary}[theorem]{Corollary}
\newtheorem{lemma}[theorem]{Lemma}
\newtheorem{claim}[theorem]{Claim}
\theoremstyle{definition}
\newtheorem{definition}[theorem]{Definition}
\theoremstyle{remark}
\newtheorem*{note*}{Note}
\renewcommand{\proofname}{\bfseries Proof}

\def\qedx{}

\newenvironment{keywords}%
  {\centerline{\bf\small Keywords}\begin{quote}\small}%
  {\par\end{quote}\vskip 1ex}

\def\equa{\smash{\stackrel{\raisebox{0.8ex}{$\scriptstyle+$}}{\smash=}}}
\def\leqa{\smash{\stackrel{\raisebox{1ex}{$\scriptstyle\!\!\;+$}}{\smash\leq}}}
\def\geqa{\smash{\stackrel{\raisebox{1ex}{$\scriptstyle+\!\!\;$}}{\smash\geq}}}
\def\eqm{\smash{\stackrel{\raisebox{0.6ex}{$\scriptstyle\times$}}{\smash=}}}
\def\leqm{\smash{\stackrel{\raisebox{1ex}{$\scriptstyle\!\!\;\times$}}{\smash\leq}}}
\def\geqm{\smash{\stackrel{\raisebox{1ex}{$\scriptstyle\times\!\!\;$}}{\smash\geq}}}

\def\nq{\hspace{-1em}}
\def\odt{{\textstyle{1\over 2}}}

\def\SetR{I\!\!R}
\def\SetN{I\!\!N}
\def\E{{\bf E}}                         
\def\M{{\cal M}}                        
\def\P{{\bf P}}                         
\def\X{{\cal X}}                        
\def\qmbox#1{{\quad\mbox{#1}\quad}}
\def\l{{\ell}}                          
\def\lb{{\log_2}}
\def\epstr{\epsilon}                    
\def\toinfty#1{\stackrel{\smash{#1}\to\infty}{\longrightarrow}}
\def\a{\alpha}
\def\o{\omega}
\def\e{{\rm e}}                        
\def\KM{K\!M}
\def\Km{K\!m}
\def\KP{K}
\def\C{C}
\def\KPM{{K_*}}
\newcommand{\kpc}[2]{\KP(#1|#2)}   
\newcommand{\kpm}[2]{\KPM(#1|#2*)} 
\newcommand{\bin}{\{0,1\}}
\newcommand{\fin}{\bin^\ast}
\newcommand{\infin}{\bin^\infty}
\newcommand{\emptyword}{\epstr} 
\newcommand*{\pair}[1]{\langle #1\rangle}

\begin{document}

\title{\normalsize
\vskip 2mm\bf\Large\hrule height5pt \vskip 6mm
Algorithmic Complexity Bounds \\
on Future Prediction Errors\thanks{
This work was supported by SNF grants
200020-107590/1,
2100-67712 and 200020-107616.
A shorter version appeared in the proceedings of the ALT'05 conference~\cite{Chernov:05postbnd}.
}
\vskip 6mm \hrule height2pt \vskip 5mm}
\author{{{\bf Alexey Chernov}$^{1,4}$
    \; {\bf Marcus Hutter}$^{1,3}$
    \; {\bf J\"{u}rgen Schmidhuber}$^{1,2}$}\\[3mm]
\normalsize $^1$IDSIA, Galleria 2, CH-6928\ Manno-Lugano, Switzerland\\
\normalsize $^2$TU Munich, Boltzmannstr.\ 3, 85748 Garching, M\"{u}nchen, Germany\\
\normalsize $^3$RSISE/ANU/NICTA, Canberra, ACT, 0200, Australia\\
\normalsize $^4$LIF, CMI, 39 rue Joliot Curie, 13453 Marseille cedex 13, France\\
\normalsize \{alexey,marcus,juergen\}@idsia.ch, \ http://www.idsia.ch/$^{_{_\sim}}\!$\{alexey,marcus,juergen\}
}
\date{19 January 2007}
\maketitle

\begin{abstract}
We bound the future loss when predicting any (computably)
stochastic sequence online. Solomonoff finitely bounded the total
deviation of his universal predictor $M$ from the true
distribution $\mu$ by the algorithmic complexity of $\mu$. Here we
assume that we are at a time $t>1$ and have already observed $x=x_1...x_t$.
We bound the future prediction performance on $x_{t+1}x_{t+2}...$
by a new variant of algorithmic complexity of $\mu$ given $x$,
plus the complexity of the randomness deficiency of $x$. The new
complexity is monotone in its condition in the sense that this
complexity can only decrease if the condition is prolonged. We
also briefly discuss potential generalizations to Bayesian model
classes and to classification problems.
\end{abstract}


\newpage
\section{Introduction}\label{secInt}

We consider the problem of online=sequential predictions. We
assume that the sequences $x=x_1x_2x_3...$ are drawn from some ``true''
but unknown probability distribution $\mu$. Bayesians proceed by
considering a class $\M$ of models=hypotheses=distributions,
sufficiently large such that $\mu\in\M$, and a prior over $\M$.
Solomonoff considered the truly large class that contains all
computable probability distributions~\cite{Solomonoff:64}.
He showed that his universal distribution $M$ converges rapidly to
$\mu$~\cite{Solomonoff:78}, i.e.\ predicts well in any environment
as long as it is computable or can be modeled by a computable
probability distribution (all physical theories are of this sort).
$M(x)$ is roughly $2^{-\KP(x)}$, where $\KP(x)$ is the length of
the shortest description of $x$, called the Kolmogorov complexity of
$x$. Since $\KP$ and $M$ are incomputable, they have to be
approximated in practice.
See e.g.~\cite{Schmidhuber:02speed,Hutter:04uaibook,Li:97,Cilibrasi:05} and
references therein.
The universality of $M$ also precludes useful statements about the
prediction quality at particular time
instances~$n$~\cite[p.\,62]{Hutter:04uaibook},
as opposed to simple classes like
i.i.d.\ sequences (data) of size $n$, where accuracy is typically
$O(n^{-1/2})$.
Luckily, bounds on the expected \emph{total}=cumulative loss
(e.g.\ number of prediction errors) for $M$ can be
derived~\cite{Solomonoff:78,Hutter:01errbnd,Hutter:02spupper,Hutter:03optisp},
which is often sufficient in an online setting. The bounds are in
terms of the (Kolmogorov) complexity of $\mu$. For instance, for
deterministic $\mu$, the number of errors is (in a sense tightly)
bounded by $\KP(\mu)$ which measures in this case the information
(in bits) in the observed infinite sequence~$x$.

\paradot{What's new}
In this paper we assume we are at a time $t>1$ and have already
observed $x=x_1...x_t$. Hence we are interested in the future
prediction performance on $x_{t+1}x_{t+2}...$, since typically we
don't care about past errors.
If the total loss is finite, the future loss must necessarily be
small for large $t$. In a sense the paper intends to quantify this
apparent triviality.
If the complexity of $\mu$ bounds the total loss, a natural guess is
that something like the conditional complexity of $\mu$ given $x$
bounds the future loss. (If $x$ contains a lot of (or even all)
information about $\mu$, we should make fewer (no) errors anymore.)
Indeed, we prove two bounds of this kind but with additional terms
describing structural properties of $x$. These additional terms
appear since the total loss is bounded only in expectation, and
hence the future loss is small only for ``most'' $x_1...x_t$. In the
first bound (Theorem~\ref{thm:lbnd}), the additional term is the
complexity of the length of $x$ (a kind of worst-case estimation).
The second bound (Theorem~\ref{thm:KPMbound}) is finer: the
additional term is the complexity of the randomness deficiency of
$x$. The advantage is that the deficiency is small for ``typical''
$x$ and bounded on average (in contrast to the length). But in this
case the conventional conditional complexity turned out to be
unsuitable.
So we introduce a new natural modification of conditional
Kolmogorov complexity, which is monotone as a function of
condition. Informally speaking, we require programs
(=descriptions) to be consistent in the sense that if a program
generates some $\mu$ given $x$, then it must generate the same
$\mu$ given any prolongation of $x$.
The new posterior bounds also significantly improve upon the
previous total bounds.

\paradot{Contents}
The paper is organized as follows. Some basic notation and
definitions are given in Sections~\ref{secNot} and~\ref{secSetup}.
In Section~\ref{secPostBnd} we prove and discuss the length-based
bound Theorem~\ref{thm:lbnd}. In Section~\ref{secKpmBnd} we show why
a new definition of complexity is necessary and formulate the
deficiency-based bound Theorem~\ref{thm:KPMbound}. We discuss the
definition and basic properties of the new complexity in
Section~\ref{secKpmDisc}, and prove Theorem~\ref{thm:KPMbound} in
Section~\ref{secKpmProof}. We briefly discuss potential
generalizations to general model classes $\M$ and classification in
the concluding Section~\ref{secDisc}.

\section{Notation \& Definitions}\label{secNot}

We essentially follow the notation of~\cite{Li:97,Hutter:04uaibook}.

\paradot{Strings and natural numbers}
We write $\X^*$ for the set of finite strings over a finite
alphabet $\X$, and $\X^\infty$ for the set of infinite sequences.
The cardinality of a set $\cal S$ is denoted by $|{\cal S}|$. We
use letters $i,k,l,n,t$ for natural numbers, $u,v,x,y,z$ for
finite strings, $\epstr$ for the empty string, and
$\a=\a_{1:\infty}$ etc.\ for infinite sequences. For a string $x$
of length $\l(x)=n$ we write $x_1x_2...x_n$ with $x_t\in\X$ and
further abbreviate $x_{k:n}:=x_kx_{k+1}...x_{n-1}x_n$ and
$x_{<n}:=x_1... x_{n-1}$. For $x_t\in\X$, denote by $\bar x_t$ an
(arbitrary) element from $\X$ such that $\bar x_t\neq x_t$. For
binary alphabet $\X=\bin$, the $\bar x_t$ is uniquely defined. We
occasionally identify strings with natural numbers.

\paradot{Prefix sets}
A string $x$ is called a (proper) prefix of $y$ if there is a
$z(\neq\epstr)$ such that $xz=y$; $y$ is called a prolongation of
$x$. We write $x*=y$ in this case, where $*$ is a wildcard for a
string, and similarly for the case where $y$ is an infinite
sequence. A set of strings is called prefix free if no element is
a proper prefix of another. Any prefix-free set $\cal P$ has the
important property of satisfying Kraft's inequality
$\sum_{x\in\cal P} |\X|^{-\l(x)}\leq 1$.

\paradot{Asymptotic notation}
We write $f(x)\leqm  g(x)$ for $f(x)=O(g(x))$ and $f(x)\leqa
g(x)$ for $f(x)\leq g(x)+O(1)$. Equalities $\eqm$, $\equa$ are
defined similarly: they hold if the corresponding inequalities
hold in both directions.

\paradot{(Semi)measures}
We call $\rho:\X^*\to[0,1]$ a semimeasure \emph{iff}
$\sum_{x_n\in\X}\rho(x_{1:n}) \le \rho(x_{<n})$ and $\rho(\epstr)
\le 1$, and a measure \emph{iff} both unstrict inequalities are
equalities. $\rho(x)$ is interpreted as the $\rho$-probability of
sampling a sequence which starts with $x$. The conditional
probability (posterior)
\beq\label{defBayes}
  \rho(y|x) \;:=\; {\rho(xy)\over\rho(x)}
\eeq
is the $\rho$-probability that a string $x$ is followed by
(continued with) $y$. If $\rho(x)=0$, $\rho(y|x)$ is defined
arbitrarily and every such function is called a version of
conditional probability.
We call $\rho$ deterministic if $\exists\a:\rho(\a_{1:n})=1$
$\forall n$. In this case we identify $\rho$ with $\a$.

\paradot{Random events and expectations}
We assume that sequence $\o=\o_{1:\infty}$ is sampled from the
``true'' measure $\mu$, i.e.\ $\P[\o_{1:n}=x_{1:n}]=\mu(x_{1:n})$.
We denote expectations w.r.t.\ $\mu$ by $\E$, i.e.\ for a function
$f:\X^n\to\SetR$,
$\E[f]=\E[f(\o_{1:n})]=\sum_{x_{1:n}}\mu(x_{1:n})f(x_{1:n})$. We
abbreviate $\mu_t:=\mu(\cdot|\o_{<t})$.

\paradot{Enumerable sets and functions}
A set of strings (or naturals, or other constructive objects) is
called \emph{enumerable} if it is the range of some computable
function. A function $f\colon\X^*\to\SetR$ is called
\emph{(co-)enumerable} if the set of pairs $\{\pair{x,\frac{k}{n}}\mid
f(x)\stackrel{\smash{\scriptscriptstyle (<)}}{\scriptstyle
>}\frac{k}{n}\}$ is enumerable.
A measure $\mu$ is called \emph{computable} if it is enumerable and
co-enumerable and the set $\{x\mid\mu(x)=0\}$ is decidable (i.\,e.\
enumerable and co-enumerable).

To simplify the statements of the theorems below, we assume that for
every computable measure $\mu$, there is one fixed computable
version of conditional probability $\mu(y|x)$, for example,
$\mu(y|x)$ is the uniform measure on $y$'s for $\mu(x)=0$.

\paradot{Prefix Kolmogorov complexity}
The conditional prefix complexity
$\kpc{y}{x}:=\min\{\l(p):U(p,x)=y\}$
is the length of the shortest binary
(self-delimiting) program $p\in\fin$ on a universal prefix Turing
machine $U$ with output $y\in\X^*$ and input $x\in\X^*$~\cite{Li:97}.
$\KP(x):=\kpc{x}{\epstr}$.
For non-string objects $o$ we define $\KP(o):=\KP(\langle
o\rangle)$, where $\langle o\rangle\in\X^*$ is some standard code
for $o$. In particular, if $(f_i)_{i=1}^\infty$ is an enumeration
of all (co-)enumerable functions, we define
$\KP(f_i):=\KP(i)$.
We need the following properties: %
The co-enumerability of $\KP$, %
the upper bounds $\kpc{x}{\l(x)}\leqa\l(x)\lb|\X|$ %
and $\KP(n)\leqa 2\lb n$, %
Kraft's inequality $\sum_x 2^{-\KP(x)}\leq 1$, %
the lower bound $\KP(x)\geq l(x)$ for ``most'' $x$ %
and $\KP(n)\toinfty{n}\infty$, %
extra information bounds $\kpc{x}{y}\leqa \KP(x)\leqa \KP(x,y)$, %
subadditivity $\KP(xy)\leqa \KP(x,y)\leqa \KP(y)+\kpc{x}{y}$, %
information non-increase $\KP(f(x))\leqa \KP(x)+\KP(f)$
for computable $f:\X^*\to\X^*$, %
and coding relative to a probability distribution (MDL):
if $P:\X^*\to[0,1]$ is enumerable and $\sum_x P(x)\leq 1$,
then
$\KP(x)\leqa -\lb P(x)+\KP(P)$.

\paradot{Monotone and Solomonoff complexity}
The monotone complexity $\Km(x):=\min\{\l(p):U(p)=x*\}$ is the
length of the shortest binary (possibly non-halting) program
$p\in\fin$ on a universal monotone Turing machine $U$ which
outputs a string starting with $x$. Solomonoff's prior
$M(x):=\sum_{p:U(p)=x*}2^{-\l(p)}=:2^{-\KM(x)}$ is the probability
that $U$ outputs a string starting with $x$ if provided with fair
coin flips on the input tape. Most complexities coincide within an
additive term $O(\log\l(x))$, e.g.\
$\kpc{x}{\l(x)}\leqa\KM(x)\leq\Km(x)\leq \KP(x)$, hence similar
relations as for $\KP$ hold.

\section{Setup}\label{secSetup}

\paradot{Convergent predictors}
We assume that $\mu$ is a ``true''\footnote{Also called
\em{objective} or \emph{aleatory} probability or \emph{chance}.}
sequence generating measure, also called an environment. If we know
the generating process $\mu$, and given past data $x_{<t}$, we can
predict the probability $\mu(x_t|x_{<t})$ of the next data item
$x_t$. Usually we do not know $\mu$, but estimate it from
$x_{<t}$. Let $\rho(x_t|x_{<t})$ be an estimated
probability\footnote{Also called \emph{subjective} or \emph{belief}
or \emph{epistemic} probability.} of $x_t$, given $x_{<t}$.
Closeness of $\rho(x_t|x_{<t})$ to $\mu(x_t|x_{<t})$ is desirable
as a goal in itself or when performing a Bayes decision $y_t$ that
has minimal $\rho$-expected loss
$l_t^\rho(x_{<t}):=\min_{y_t}\sum_{x_t}\mbox{Loss}(x_t,y_t)\rho(x_t|x_{<t})$.
Consider, for instance, a weather data sequence $x_{1:n}$ with
$x_t=1$ meaning rain and $x_t=0$ meaning sun at day $t$. Given
$x_{<t}$ the probability of rain tomorrow is $\mu(1|x_{<t})$. A
weather forecaster may announce the probability of rain to be
$y_t:=\rho(1|x_{<t})$, which should be close to
the true probability $\mu(1|x_{<t})$.
To aim for
\beqn\label{eqconv}
 \rho(x'_t|x_{<t}) - \mu(x'_t|x_{<t}) \;\stackrel{(fast)}\longrightarrow\; 0
 \qmbox{as} t\to\infty
\eeqn
seems reasonable.

\paradot{Convergence in mean sum}
We can quantify the deviation of $\rho_t$ from $\mu_t$, e.g.\ by
the squared difference
\beqn
  s_t(\o_{<t}) \;:=\; \sum_{x_t\in\X}(\rho(x_t|\o_{<t})-\mu(x_t|\o_{<t}))^2
  \;\equiv\;\sum_{x_t}(\rho_t-\mu_t)^2
\eeqn
Alternatively one may also use the squared absolute distance
$s_t:=\odt(\sum_{x_t}|\rho_t-\mu_t|)^2$, the Hellinger distance
$s_t:=\sum_{x_t}(\sqrt{\rho_t}-\sqrt{\mu_t})^2$, the KL-divergence
$s_t:=\sum_{x_t}\mu_t\ln{\mu_t\over\rho_t}$, or the squared Bayes
regret $s_t:=\odt(l_t^\rho-l_t^\mu)^2$ for $l_t\in[0,1]$. For all these
distances one can show~\cite{Hutter:01loss,Hutter:02spupper,Hutter:04uaibook}
that their cumulative expectation from $l$ to $n$ is bounded as follows:
\beq\label{eq:KL}
  0 \;\leq\; \E\left[\left.\sum_{t=l}^n s_t\right|\o_{<l}\right]
  \;\leq\; \E\left[\left.\ln{\mu(\o_{l:n}|\o_{<l})\over\rho(\o_{l:n}|\o_{<l})}
             \right|\o_{<l}\right]
  \;=:\; D_{l:n}(\o_{<l}).
\eeq
$D_{l:n}$ is increasing in $n$, hence $D_{l:\infty}\in[0,\infty]$
exists~\cite{Hutter:01alpha,Hutter:04uaibook}.
A sequence of random variables like $s_t$ is said to converge to
zero with probability 1 if the set $\{\omega :
s_t(\omega)\,\toinfty{t}\,0\}$ has measure 1. $s_t$ is said to
converge to zero in mean sum if $\sum_{t=1}^\infty\E\bigl[|s_t|\bigr]\leq
c<\infty$, which implies convergence with probability 1 (rapid if
$c$ is of reasonable size).
Therefore a small finite bound on $D_{1:\infty}$ would imply rapid
convergence of the $s_t$ defined above to zero, hence
$\rho_t\to\mu_t$ and $l_t^\rho\to l_t^\mu$ fast. So the crucial
quantities to consider and bound (in expectation) are
$\ln\smash{\mu(x)\over\rho(x)}$ if $l=1$ and
$\ln\smash{\mu(y|x)\over\rho(y|x)}$ for $l>1$. For illustration we will
sometimes loosely interpret $D_{1:\infty}$ and other quantities as
the number of prediction errors, as for the error-loss they are
closely related to it~\cite{Hutter:01errbnd,Hutter:01alpha}.

\paradot{Bayes mixtures}
A Bayesian considers a class of distributions $\M :=
\{\nu_1,\nu_2,...\}$, large enough to contain $\mu$, and uses
the Bayes mixture
\beq\label{xidefsp}
  \xi(x) \;:=\;
  \sum_{\nu\in\M}w_\nu\!\cdot\!\nu(x),\quad
  \sum_{\nu\in\M}w_\nu=1,\quad w_\nu>0
\eeq
for prediction, where $w_\nu$ can be interpreted as the prior of
(or initial belief in) $\nu$. The dominance
\beq\label{unixi}
  \xi(x) \;\geq\;
  w_\mu\!\cdot\!\mu(x) \quad \forall x\in\X^*
\eeq
is its most important property. Using $\rho=\xi$ for prediction,
this implies $D_{1:\infty}\leq\ln w_\mu^{-1}<\infty$, hence
$\xi_t\to\mu_t$. If $\M$ is chosen sufficiently large, then
$\mu\in\M$ is not a serious constraint.

\paradot{Solomonoff prior}
So we consider the largest (from a computational point of view)
relevant class, the class $\M_U$ of all enumerable semimeasures
(which includes all computable probability distributions) and
choose $w_\nu=2^{-\KP(\nu)}$ which is biased towards simple
environments (Occam's razor). This gives us Solomonoff-Levin's
prior $M$~\cite{Solomonoff:64,Zvonkin:70} (this definition
coincides within an irrelevant multiplicative constant with the
one in Section~\ref{secNot}). In the following we assume
$\M=\M_U$, $\rho=\xi=M$, $w_\nu=2^{-\KP(\nu)}$ and $\mu\in\M_U$
being a computable (proper) measure, hence $M(x)\geq
2^{-K(\mu)}\mu(x)\,\forall x$ by~(\ref{unixi}).

\paradot{Prediction of deterministic environments}
Consider a computable sequence $\a=\a_{1:\infty}$ ``sampled from
$\mu\in\M$'' with $\mu(\a)=1$, i.e.\ $\mu$ is deterministic,
then from~(\ref{unixi}) we get
\beq\label{detBnd}
  \sum_{t=1}^\infty |1-M(\a_t|\a_{<t})|
  \;\leq\; - \sum_{t=1}^\infty \ln M(\a_t|\a_{<t})
  \;=\; -\ln M(\a_{1:\infty})
  \;\leq\; \KP(\mu) \ln 2< \infty,
\eeq
which implies that $M(\a_t|\a_{<t})$ converges rapidly to 1 and
hence $M(\bar\a_t|\a_{<t})\to 0$, i.e.\ asymptotically $M$
correctly predicts the next symbol. The number of prediction
errors is of the order of the complexity $\KP(\mu)\equa\Km(\a)$ of the
sequence.

For binary alphabet this is the best we can expect, since at each
time-step only a single bit can be learned about the environment,
and only after we ``know'' the environment we can predict
correctly. For non-binary alphabet, $\KP(\mu)$ still measures the
information in $\mu$ in bits, but feedback per step can now be $\lb|\X|$
bits, so we may expect a better bound $\KP(\mu)/\lb|\X|$. But in the worst
case all $\a_t\in\{0,1\}\subseteq\X$. So without structural
assumptions on $\mu$ the bound cannot be improved even if $\X$ is
huge. We will see how our posterior bounds can help in this
situation.

\paradot{Individual randomness (deficiency)}
Let us now consider a general (not necessarily deterministic)
computable measure $\mu\in\M$. The Shannon-Fano code of $x$ w.r.t.\ $\mu$
has code-length $\lceil-\lb\mu(x)\rceil$, which is ``optimal''
for ``typical/random'' $x$ sampled from $\mu$. Further, $-\lb
M(x)\approx \KP(x)$ is the length of an ``optimal'' code for $x$.
Hence $-\lb\mu(x)\approx -\lb M(x)$ for ``$\mu$-typical/random''
$x$. This motivates the definition of \emph{$\mu$-randomness deficiency}
\beqn
  d_\mu(x) \;:=\; \lb{M(x)\over\mu(x)}
\eeqn
which is small for ``typical/random'' $x$.
Formally, a sequence $\a$ is called (Martin-L\"of)
random iff
$d_\mu(\a):=\sup_n d_\mu(\a_{1:n})<\infty$, i.e.\ iff its
Shannon-Fano code is ``optimal'' (note that
$d_\mu(\a)\geq-\KP(\mu)>-\infty$ for all sequences), i.e.\ iff
\beqn
  \sup_n\Big|\sum_{t=1}^n\log{\mu(\a_t|\a_{<t})\over M(\a_t|\a_{<t})}\Big|
  \;\equiv\; \sup_n\Big|\log{\mu(\a_{1:n})\over M(\a_{1:n})}\Big|
  \;<\; \infty.
\eeqn
Unfortunately this does not imply $M_t\to\mu_t$ on the
$\mu$-random $\a$, since $M_t$ may oscillate around $\mu_t$, which
indeed can happen~\cite{Hutter:04mlconvx}. But if we take the
expectation,
Solomonoff~\cite{Solomonoff:78,Hutter:01alpha,Hutter:04uaibook} showed
\beq\label{prBnd}
  0 \;\leq\; \sum_{t=1}^\infty \E\sum_{x_t}(M_t-\mu_t)^2
  \;\leq\; D_{1:\infty}
  \;=\; \lim_{n\to\infty}\E[-d_\mu(\o_{1:n})]\ln 2
  \;\leq\; \KP(\mu)\ln 2
  \;<\; \infty\,,
\eeq
hence, $M_t\to\mu_t$ with $\mu$-probability 1. So in any case,
$d_\mu(x)$ is an important quantity, since the smaller $-d_\mu(x)$
(at least in expectation) the better $M$ predicts.

\section{Posterior Bounds}\label{secPostBnd}

\paradot{Posterior bounds}
Both bounds~(\ref{detBnd}) and~(\ref{prBnd}) bound the total
(cumulative) discrepancy (error) between $M_t$ and $\mu_t$. Since
the discrepancy sum $D_{1:\infty}$ is finite, we know that after sufficiently
long time $t=l$, we will make few further errors, i.e.\ the
future error sum $D_{l:\infty}$ is small.
The main goal of this paper is to quantify this asymptotic
statement. So we need bounds on $\lb\smash{\mu(y|x)\over M(y|x)}$, where
$x$ are the past and $y$ the future observations. Since
$\lb\smash{\mu(y)\over M(y)}\leq \KP(\mu)$ and $\mu(y|x)/M(y|x)$ are
conditional versions of true/universal distributions, it seems
natural that the unconditional bound $\KP(\mu)$ also simply
conditionalizes to $\lb\smash{\mu(y|x)\over M(y|x)}\leq\kpc{\mu}{x}$.
The more information the past observation $x$ contains about $\mu$,
the easier it is to code $\mu$ i.e.\ the smaller $\kpc{\mu}{x}$ is,
and hence the less future predictions errors $D_{l:\infty}$ we
should make. Once $x$ contains all information about $\mu$, i.e.\
$\kpc{\mu}{x}\equa 0$, we should make no errors anymore.
More formally, optimally coding $x$, then $\mu|x$, and finally
$y|\mu,x$ by Shannon-Fano gives a code for $xy$, hence
$\KP(xy)\lesssim \KP(x)+\kpc{\mu}{x}-\lb\mu(y|x)$. Since
$\KP(z)\approx-\lb M(z)$, this implies $\lb{\mu(y|x)\over
M(y|x)}\lesssim \kpc{\mu}{x}$, but with a logarithmic fudge that
tends to infinity as $\l(y)\to\infty$, which is unacceptable. The
$y$-independent bound we need was first stated
in~\cite[Prob.2.6$(iii)$]{Hutter:04uaibook}:

\begin{theorem}\label{thm:lbnd}
For any computable measure $\mu$ and any $x,y\in\X^*$ it holds that
\beqn
  \lb\frac{\mu(y|x)}{M(y|x)}
  \;\leqa\; \kpc{\mu}{x} + \KP(\l(x)).
\eeqn
\end{theorem}

\begin{proof}
For every $l$ we define the following function of $z\in\X^*$.
For $\l(z)\geq l$,
\beqn
  \psi^l(z) \;:=\; \sum_{\nu\in\M} 2^{-\kpc{\nu}{z_{1:l}}}M(z_{1:l})\nu(z_{l+1:\l(z)})\,.
\eeqn
For $\l(z)<l$, we extend $\psi^l$ by defining
$\psi^l(z):=\sum_{u:\l(u)=l-\l(z)}\psi^l(zu)$. It is easy to see
that $\psi^l$ is an enumerable semimeasure. By the definition of $M$,
we have
\beqn
  M(z) \;\ge\; 2^{-\KP(\psi^l)}\psi^l(z)
\eeqn
for any $l$ and $z$.
Now let $l=\l(x)$ and $z=xy$. Let us define a computable measure
$\mu^x(y):=\mu(y|x)$. Then
\beqn
  M(xy) \;\geq\; 2^{-\KP(\psi^l)}\psi^l(xy)
  \;\geq\; 2^{-\KP(\psi^l)}2^{-\kpc{\mu^x}{x}}M(x)\mu^x(y)\,.
\eeqn
Taking the logarithm, after trivial transformations, we get
\beqn
  \lb\frac{\mu(y|x)}{M(y|x)} \;\leq\;
      \kpc{\mu^x}{x} + \KP(\psi^l)\,.
\eeqn
To complete the proof, let us note that $\KP(\psi^l)\leqa\KP(l)$
and $\kpc{\mu^x}{x}\leqa\kpc{\mu}{x}$.
\qedx\end{proof}

\begin{corollary}\label{cor:lbnd}
The future deviation of $M_t$ from $\mu_t$ is bounded by
$$
   \textstyle\sum_{t=l+1}^\infty \E[s_t|\o_{1:l}]
  \;\leq\; D_{l+1:\infty}(\o_{1:l})
  \;\leqa\; (\kpc{\mu}{\o_{1:l}}\!+\!\KP(l))\ln 2
\eqno(i)
$$
For $s_t$ being squared (absolute) distance, Hellinger distance, or
squared Bayes regret,
the total deviation of $M_t$ from $\mu_t$ is bounded by
$$
   \textstyle\sum_{t=1}^\infty \E[s_t]
  \;\leqa\; \min_l\{\E[\kpc{\mu}{\o_{1:l}}\!+\!\KP(l)]\ln 2 + 2l\}
\eqno(ii)
$$
\end{corollary}

\begin{proof}
$(i)$ The first inequality is~(\ref{eq:KL}) and the second follows
by taking the conditional expectation $\E[\cdot|\o_{1:l}]$ in
Theorem~\ref{thm:lbnd}.
$(ii)$ follows from $(i)$ by taking the unconditional expectation and
from $\sum_{t=1}^l\E[s_t]\leq 2l$, since $s_t\leq 2$
for these distances~\cite{Hutter:04uaibook}.
\qedx\end{proof}

\paradot{Examples and more motivation}
The bounds Theorem~\ref{thm:lbnd} and Corollary~\ref{cor:lbnd}$(i)$
prove and quantify the intuition that the more
we know about the environment, the better our predictions. We show
the usefulness of the new bounds for some deterministic
environments $\mu\widehat=\a$.

Assume all observations are identical, i.e.\ $\a=x_1x_1x_1...$.
Further assume that $\X$ is huge and $\KP(x_1)=\lb|\X|$, i.e.\
$x_1$ is a typical/random/complex element of $\X$. For instance if
$x_1$ is a $256^3$ color 512$\times$512 pixel image, then
$|\X|=256^{3\times 512\times 512}$. Hence the standard bound~(\ref{prBnd})
 on the number of errors $D_{1:\infty}/\ln 2\leq
\KP(\mu)\equa \KP(x_1)=3\cdot 2^{21}$ is huge. Of course,
interesting pictures are not purely random, but their complexity is
often only a factor 10..100 less, so still large. On the other
hand, any reasonable prediction scheme observing a few (rather
than several thousands) identical images, should predict that the
next image will be the same. This is what our posterior bound
gives, $D_{2:\infty}(x_1)\leqa (\kpc{\mu}{x_1}+\KP(1))\ln 2\equa 0$,
hence indeed $M$ makes only $\sum_{t=1}^\infty \E[s_t]=O(1)$
errors by Corollary~\ref{cor:lbnd}$(ii)$, significantly improving
upon Solomonoff's bound $\KP(x_1)\ln 2$.

More generally, assume $\a=x\o$, where the initial part
$x=x_{1:l}$ contains all information about the remainder, i.e.\
$\kpc{\mu}{x}\equa \kpc{\o}{x}\equa 0$. For instance, $x$ may be a binary
program for $\pi$ or $\e$, and $\o$ its $|\X|$-ary expansion.
Sure, given the algorithm for some number sequence,
it should be perfectly predictable. Indeed, Theorem~\ref{thm:lbnd}
implies $D_{l+1:\infty}\leqa \KP(l)$, which can be exponentially
smaller than Solomonoff's bound $\KP(\mu)$ ($\equa l$ if $\KP(x)\equa\l(x)$). On
the other hand, $\KP(l)\geq\lb l$ for most $l$, i.e.\ is larger than
the $O(1)$ that one might hope for.

\paradot{Logarithmic versus constant accuracy}
Thus there is one blemish in the bound. There is an additive
correction of logarithmic size in the length of $x$. Many theorems
in algorithmic information theory hold to within an additive
constant, sometimes this is easily reached, sometimes with difficulty,
sometimes one needs a suitable complexity variant, and sometimes
the logarithmic accuracy cannot be improved~\cite{Li:97}. The
latter is the case with Theorem~\ref{thm:lbnd}:

\begin{lemma}\label{lem:lengthtight}
For $\X=\bin$,
for any positive computable measure $\mu$,
there exists a computable sequence $\a\in\infin$ such that for any $l\in\SetN$
\beqn
  D_{l:\infty}(\a_{<l})
  \;\geq\; D_{l:l}(\a_{<l})
  \;\equiv\; \sum_{b\in\bin}\mu(b|\a_{<l})\ln\frac{\mu(b|\a_{<l})}{M(b|\a_{<l})}
  \;\geqa\; \textstyle{1\over 3} \KP(l)\,.
\eeqn
\end{lemma}

\begin{proof}
Let us construct such a computable sequence
$\a\in\bin^\infty$ by induction. Assume that $\a_{<l}$ is
constructed. Since $\mu$ is a measure, either
$\mu(0|\a_{<l})>c$ or $\mu(1|\a_{<l})>c$ for
$c:=[3\ln 2]^{-1}<\odt$.
Since $\mu$ is computable, we can find
(effectively) $b\in\bin$ such that $\mu(b|\a_{<l})>c$.
Put
$\a_l=\bar b$.

Let us estimate $M(\bar \a_l|\a_{<l})$. Since
$\a$ is computable, $M(\a_{<l})\geqm 1$. We claim that
$M(\a_{<l}\bar \a_l)\leqm 2^{-\KP(l)}$. Actually,
consider the set $\{\a_{<l}\bar \a_l\mid l> 0\}$.
This set is prefix free and decidable. Therefore
$P(l)=M(\a_{<l}\bar \a_l)$ is an enumerable function
with $\sum_l P(l)\le 1$, and the claim follows from the coding
theorem. Thus, we have $M(\bar \a_l|\a_{<l})\leqm
2^{-\KP(l)}$ for any $l$.
Since $\mu(\bar \a_l|\a_{<l})>c$, we get
\bqan
  \sum_{b\in\bin}\!\!
  \mu(b|\a_{<l})\ln\frac{\mu(b|\a_{<l})}{M(b|\a_{<l})}
& \;\geqa\; &
  \mu(\bar\a_l|\a_{<l})\ln{c\over 2^{-\KP(l)}} +\!\!\!
  \min_{p\in[0,1-c]}p\ln\frac{p}{M(\a_l|\a_{<l})}
\\
& \;\geqa\; &
  c\KP(l)\ln 2
\eqan
\qedx\end{proof}

A constant fudge is generally preferable to a logarithmic one for
quantitative and aesthetical reasons. It also often leads to
particular insight and/or interesting new complexity variants
(which will be the case here). Though most complexity variants
coincide within logarithmic accuracy
(see~\cite{Schmidhuber:00toe,Schmidhuber:02gtm} for exceptions), they
can have very different other properties. For instance, Solomonoff
complexity $\KM(x)=-\lb M(x)$ is an excellent predictor, but
monotone complexity $\Km$ can be exponentially worse and prefix
complexity $\KP$ fails
completely~\cite{Hutter:03unimdl,Hutter:06unimdlx}.

\paradot{Exponential bounds}
Bayes is often approximated by MAP or MDL. In our context this
means approximating $\KM$ by $\Km$ with exponentially worse bounds
(in deterministic environments)~\cite{Hutter:03unimdl}.
(Intuitively, since an error with Bayes eliminates half of the
environments, while MAP/MDL may eliminate only one.) Also for more
complex ``reinforcement'' learning problems, bounds can be
$2^{\KP(\mu)}$ rather than $\KP(\mu)$ due to sparser feedback. For
instance, for a sequence $x_1x_1x_1...$ if we do not observe $x_1$
but only receive a reward if our prediction was correct, then the
only way a universal predictor can find $x_1$ is by trying out all
$|\X|$ possibilities and making (in the worst case) $|\X|-1\eqm
2^{\KP(\mu)}$ errors. Posterization allows us to boost such gross
bounds to useful bounds $2^{\kpc{\mu}{x_1}}=O(1)$. But in general,
additive logarithmic corrections as in Theorem~\ref{thm:lbnd} also
exponentiate and lead to bounds polynomial in $l$ which may be
quite sizeable. Here the advantage of a constant correction
becomes even more apparent~\cite[Problems 2.6, 3.13, 6.3 and
Section 5.3.3]{Hutter:04uaibook}.

\section{More Bounds and New Complexity Measure}\label{secKpmBnd}

Lemma~\ref{lem:lengthtight} shows that the bound in
Theorem~\ref{thm:lbnd} is attained for some binary strings. But
for other binary strings the bound may be very rough. (Similarly,
$\KP(x)$ is greater than $\l(x)$ infinitely often, but
$\KP(x)\ll\l(x)$ for many ``interesting'' $x$.) Let us try to find
a new bound, which does not depend on $\l(x)$.

First observe that, in contrast to the unconditional case~(\ref{prBnd}),
$\KP(\mu)$ is not an upper bound (again by
Lemma~\ref{lem:lengthtight}). Informally speaking, the reason is
that $M$ can predict the future very badly if the past is not
``typical'' for the environment (such past $x$ have low
$\mu$-probability, therefore in the unconditional case their
contribution to the expected loss is small). So, it is natural to
bound the loss in terms of randomness deficiency $d_\mu(x)$, which
is a quantitative measure of ``typicalness''.

\begin{theorem}\label{thm:bounddefect}
For any computable measure $\mu$ and any $x,y\in\fin$ it holds
\beqn
  \lb\frac{\mu(y|x)}{M(y|x)}
  \;\equiv\; d_\mu(x)-d_\mu(xy)
  \;\leqa\; \KP(\mu)+\KP(\lceil d_{\mu}(x)\rceil)\,.
\eeqn
\end{theorem}

Theorem~\ref{thm:bounddefect} is a variant of the
``deficiency conservation theorem'' from ~\cite{VerShen:book}.
We do not know who was the first to discover this statement
and whether it was published
(the special case where $\mu$ is the uniform measure was proved
by An.~Muchnik as an auxiliary lemma for one of his unpublished results;
then A.~Shen placed a generalized statement to the (unfinished)
book~\cite{VerShen:book}).

Now, our goal is to replace $\KP(\mu)$ in the last bound
by a conditional complexity of $\mu$.
Unfortunately, the conventional conditional prefix complexity
is not suitable:

\begin{lemma}\label{lem:defecttight}
Let $\X=\bin$.
There is a constant $C_0$ such that for any $l\in\SetN$,
there are a computable measure $\mu$ and $x\in\bin^l$ such that
\beqn
\kpc{\mu}{x}\le C_0,\quad |d_\mu(x)|\le C_0,\quad\text{and}\qquad
\eeqn
\beqn
  D_{l+1:l+1}(x) \;\equiv\;
  \sum_{b\in\bin}\mu(b|x)\ln\frac{\mu(b|x)}{M(b|x)} \;\geqa\; \KP(l)\ln 2\,.
\eeqn
\end{lemma}
\begin{proof}
For $l\in\SetN$, define a deterministic measure
$\mu^l$ such that
$\mu^l$
is equal to $1$ on the prefixes of $0^l1^\infty$
and is equal to $0$ otherwise.

Let $x=0^l$. Then $\mu^l(x)=1$, $\mu^l(x0)=0$, $\mu^l(x1)=1$.
Also $1\ge M(x)\ge M(x0)\ge M(0^\infty)\eqm 1$
and (as in the proof of Lemma~\ref{lem:lengthtight})
$M(x1)\leqm 2^{-\KP(l)}$.
Trivially, $d_{\mu^l}(x)=\lb M(x) \eqm 1$,
and $\kpc{\mu^l}{x}\equa\kpc{\mu^l}{l}\equa 0$.
Thus, $\kpc{\mu^l}{x}$ and $d_{\mu^l}(x)$ are bounded
by a constant $C_0$ independent of $l$.
On the other hand,
\beqn
 \sum_{b\in\bin}\mu^l(b|x)\ln\frac{\mu^l(b|x)}{M(b|x)}
 \;=\; \ln\frac{1}{M(1|x)} \;\geqa\; \KP(l)\ln 2\,.
\eeqn
(One can obtain the same result also for non-deterministic $\mu$,
for example, taking $\mu^l$ mixed with the uniform measure.)
\qedx\end{proof}

Informally speaking, in Lemma~\ref{lem:defecttight} we exploit the
fact that $\kpc{y}{x}$ can use the information about the length of
the condition $x$. Hence $\kpc{y}{x}$ can be small for a certain
$x$ and is large for some (actually almost all) prolongations of
$x$. But in our case of sequence prediction, the length of $x$
grows taking all intermediate values and cannot contain any
relevant information. Thus we need a new kind of conditional
complexity.

Consider a Turing machine $T$ with two input tapes. Inputs are
provided without delimiters, so the size of the input is defined by
the machine itself. Let us call such a machine \emph{twice
prefix}. We write that $T(x,y)=z$ if machine $T$, given a sequence
beginning with $x$ on the first tape and a sequence beginning with
$y$ on the second tape, halts after reading exactly $x$ and $y$
and prints $z$ to the output tape. (Obviously, if $T(x,y)=z$, then
the computation does not depend on the contents of the input tapes
after $x$ and $y$.) We define
\beqn
  \C_T(y|x) \;:=\; \min\{\l(p)\mid\exists k\le\l(x) :\: T(p,x_{1:k})=y\}\,.
\eeqn
Clearly, $\C_T(y|x)$ is an enumerable from above function of $T$,
$x$, and $y$. Using a standard argument~\cite{Li:97}, one can show
that there exists an optimal twice prefix machine $U$ in the sense
that for any twice prefix machine $T$
\beqn
  \C_U(y|x) \;\leqa\; \C_T(y|x)\,.
\eeqn

\begin{definition}
\emph{Complexity monotone in conditions} is defined for some fixed
optimal twice prefix machine $U$ as
\beqn
  \kpm{y}{x} \;:=\; \C_U(y|x)
  \;=\; \min\{\l(p)\mid\exists k\le\l(x) : U(p,x_{1:k})=y\}\,.
\eeqn
\end{definition}
\noindent
Here $*$ in $x*$ is a syntactical part of the complexity notation
$\kpm{\cdot}{\cdot}$,
though one may think of $\kpm{y}{x}$ as of the minimal length of
a program that produces $y$ given any $z=x*$.

\begin{theorem}\label{thm:KPMbound}
For any computable measure $\mu$ and any $x,y\in\X^*$ it holds
\beqn
  \lb\frac{\mu(y|x)}{M(y|x)}
  \;\leqa\; \kpm{\mu}{x}+\KP(\lceil d_{\mu}(x)\rceil)\,.
\eeqn
\end{theorem}

\begin{note*}
One can get slightly stronger variants of Theorems~\ref{thm:lbnd}
and~\ref{thm:KPMbound}
by replacing the complexity of a standard code of $\mu$
by more sophisticated values.
First, in any effective encoding there are many codes for every $\mu$,
and in all the upper bounds (including Solomonoff's one)
one can take the minimum of the complexities
of all the codes for $\mu$.
Moreover, in Theorem~\ref{thm:lbnd} it is sufficient
to take the complexity of $\mu^x=\mu(\cdot|x)$
(and it is sufficient that $\mu^x$ is enumerable,
while $\mu$ can be incomputable).
For Theorem~\ref{thm:KPMbound} one can prove a similar strengthening:
The complexity of $\mu$ is replaced by the complexity
of any computable function
that is equal to $\mu$ on all prefixes and prolongations of~$x$.
\end{note*}

To demonstrate the usefulness of the new bound, let us again
consider some deterministic environment $\mu\widehat=\a$.
For $\X=\bin$ and $\a=x^\infty$ with $x=0^n1$,
Theorem~\ref{thm:lbnd} gives
the bound ${\kpc{\mu}{n}+\KP(n)}\equa\KP(n)$.
Consider the new bound
$\kpm{\mu}{x}+\KP(\lceil d_\mu(x)\rceil)$.
Since $\mu$ is deterministic,
we have $d_\mu(x)=\lb M(x)\equa -\KP(n)$, and
$\KP(\lceil d_\mu(x)\rceil)\equa\KP(\KP(n))$.
To estimate $\kpm{\mu}{x}$,
let us consider a machine $T$ that reads
only its second tape and outputs the number of $0$s before the
first $1$. Clearly, $\C_T(n|x)=0$, hence $\kpm{\mu}{x}\equa 0$.
Finally, $\kpm{\mu}{x}+\KP(\lceil d_\mu(x)\rceil)\leqa
\KP(\KP(n))$, which is much smaller than $\KP(n)$.

\section{Properties of the New Complexity}\label{secKpmDisc}

The above definition of $\KPM$ is based on computations of some
Turing machine. Such definitions are quite visual, but are often
not convenient for formal proofs. We will give an alternative
definition in terms of enumerable sets (see~\cite{UspShen:96} for
definitions of unconditional complexities in this style), which
summarizes the properties we actually need for the proof of
Theorem~\ref{thm:KPMbound}.

An enumerable set $E$ of triples of strings is called
\emph{$\KPM$-correct} if it satisfies the following requirements:
\begin{enumerate}%
\item if $\pair{p,x,y_1}\in E$ and $\pair{p,x,y_2}\in E$,
 then $y_1=y_2$;
\item if $\pair{p,x,y}\in E$, then $\pair{p',x',y}\in E$
for all $p'$ being prolongations of $p$
and all $x'$ being prolongations of $x$;
\item if $\pair{p,x',y}\in E$ and $\pair{p',x,y}\in E$,
and $p$ is a prefix of $p'$ and $x$ is a prefix of $x'$,
then $\pair{p,x,y}\in E$.
\end{enumerate}
A complexity of $y$ under a condition $x$ w.r.t. a set $E$ is
\beqn
  \C_E(y|x) \;=\; \min\{\l(p)\mid\pair{p,x,y}\in E\}\,.
\eeqn
A $\KPM$-correct set $E$ is called \emph{optimal} if
\beqn
  \C_{E}(y|x)\leqa\C_{E'}(y|x)
\eeqn
for any $\KPM$-correct set $E'$.
One can easily construct an enumeration of all $\KPM$-correct
sets, and an optimal set exists by the standard argument.

It is easy to see that a twice prefix Turing machine $T$ can be
transformed to a set $E$ such that $\C_T(y|x)=\C_E(y|x)$. The set
$E$ is constructed as follows: $T$ is run on all possible inputs,
and if $T(p,x)=y$, then pairs $\pair{p',x',y}$ are added to $E$
for all $p'$ being prolongations of $p$ and all $x'$ being
prolongations of $x$. Evidently, $E$ is enumerable, and the second
requirement of $\KPM$-correctness is satisfied. To verify the
other requirements, let us consider arbitrary
$\pair{p'_1,x'_1,y_1}\in E$ and $\pair{p'_2,x'_2,y_2}\in E$ such
that $p'_1$ and $p'_2$, $x'_1$ and $x'_2$ are comparable (one is a
prefix of the other).
By construction of $E$,
there are $p_i$ being prefixes of $p'_i$
and $x_i$ being prefixes of $x'_i$  such that
$T(p_i,x_i)=y_i$ for $i=1,2$.
Clearly, $p_1$ and $p_2$,
$x_1$ and $x_2$ are comparable too.
Since replacing the unused part
of the inputs does not affect the running of the machine $T$ and
comparable words have a common prolongation, we get $p_1=p_2$,
$x_1=x_2$, and $y_1=y_2$. Thus $E$ is a $\KPM$-correct set.

The transformation in the other direction is impossible in some cases:
the set $E=\{\pair{0^{h(n)}p,0^n1q,0}\mid n\in\SetN,\: p,q\in\fin\}$,
where $h(n)$ is $0$ if the $n$-th Turing machine halts and $1$
otherwise, is $\KPM$-correct, but does not have a corresponding
machine $T$.
(Assume that such a machine $T$ exists.
If the $n$-th machine halts,
then $\pair{\epstr,0^n1,0}\in E$ and thus $T$ does not read
the input tape at all.
If the $n$-th machine does not halt,
then $\pair{0,0^n1,0}\in E$ and $\pair{1,0^n1,0}\notin E$ and thus
$T$ has to read first symbol on the input tape.
Therefore, one can use $T$ to solve the halting problem.)
However, we conjecture (but cannot prove) that for every set $E$ there
exists a machine $T$ such that $\C_T(x|y)\equa\C_E(x|y)$.

Probably, the requirements on $E$ can be even weaker, namely, the
third requirement might be superfluous. Let us notice that the first
requirement of $\KPM$-correctness allows us to consider
the set $E$ as a partial computable function:
$E(p,x)=y$ iff $\pair{p,x,y}\in E$. The
second requirement says that $E$ becomes a continuous function if
we take the topology of prolongations (any neighborhood of
$\pair{p,x}$ contains the cone $\{\pair{p*,x*}\}$) on the
arguments and the discrete topology ($\{y\}$ is a neighborhood of
$y$) on values. It is known (see~\cite{UspShen:96} for references)
that different complexities (plain, prefix, decision) can be
naturally defined in a similar ``topological'' fashion. We
conjecture the same is true in our case: an optimal enumerable set
satisfying the requirements (1) and (2) (obviously, it exists)
specifies the same complexity (up to an additive constant) as an
optimal twice prefix machine.

It follows immediately from the definition(s) that
$\kpm{y}{x}$ is monotone as a function of $x$:
$\kpm{y}{xz}\leq\kpm{y}{x}$ for all $x$, $y$, $z$.

The following lemma provides bounds
for $\kpm{x}{y}$ in terms of prefix complexity $\KP$.
The lemma holds for all the definitions of $\kpm{x}{y}$ above.

\begin{lemma}\label{lem:KPMK}
For any $x,y\in\X^*$ it holds
\beqn
\kpc{x}{y} \;\leqa\;  \kpm{x}{y} \;\leqa\;
  \min_{l\le\l(y)}\{\kpc{x}{y_{1:l}}+\KP(l)\} \;\leqa\; \KP(x)\,.
\eeqn
In general, none of the bounds are equal to $\kpm{x}{y}$ even within
a $o(\KP(x))$ term, but they are attained for certain $y$: For every
$x$ there is a $y$ such that
\beqn
  \kpc{x}{y} \;\equa\; 0 \qquad\text{and}\qquad
  \kpm{x}{y} \;\equa\; \min_{l\le\l(y)}\{\kpc{x}{y_{1:l}}+\KP(l)\}\;\equa\; \KP(x) \,,
\eeqn
and for every $x$ there is a $y$ such that
\beqn
  \kpc{x}{y} \;\equa\;\kpm{x}{y}\equa 0 \qquad\text{and}\qquad
  \min_{l\le\l(y)}\{\kpc{x}{y_{1:l}}+\KP(l)\} \;\equa\; \KP(x)\,.
\eeqn
\end{lemma}

\begin{proof}
The first inequality is trivial
(any twice-prefix machine is also a prefix machine in the first argument),
as well as the last one (consider $l=0$).
Let us describe a twice prefix machine
that provides $\kpm{x}{y}\leqa \min_{l\le\l(y)}\{\kpc{x}{y_{1:l}}+\KP(l)\}$.
The first tape contains a prefix code $p_l$ of $l$
followed by a prefix code $p$ for $x$ under condition $y_{1:l}$,
and the second tape contains $y$.
The machine reads the $p_l$ on the first tape
and reconstructs the number $l$,
then reads $l$ bits from the second tape,
and then reads $p$ using these bits as the condition.
Thus, $\kpm{x}{y}\leqa\l(p_l)+\l(p)\leqa
\KP(l)+\kpc{x}{y_{1:l}}$.

Let us show that the bounds are attained.

Let us observe that $\KP(x)\leqa\kpm{x}{0^n}$ for all $x$ and $n$.
Actually, let
$P(x)=\max\{2^{-\l(p)}\mid \exists n\:\pair{p,0^n,x}\in E\}$
(which implies $-\lb P(x)\le\kpm{x}{0^n}$ for all $n$).
Obviously, $P(x)$ is enumerable.
Further, $\sum_x P(x)\le 1$ since
$\sum_x P(x)$ is a sum of $2^{-\l(p)}$
over a prefix-free set of $p$
(Assume the converse, $p$ is a prefix of $q$,
and $\pair{p,0^n,x}\in E$, $\pair{q,0^m,y}\in E$
for some $n$, $m$, and different $x$, $y$.
By the second requirement of $\KPM$-correctness,
$\pair{q,0^{\max\{m,n\}},x}\in E$,
$\pair{q,0^{\max\{m,n\}},y}\in E$.
By the first requirement, $x=y$, contradiction.)
Thus, by the coding theorem,
$\KP(x)\leqa -\lb P(x)\leqa\kpm{x}{0^n}$.

To get the first example,
for arbitrary $x$, let us take $y=0^n$ such that $n$
is the number of $x$ in some ordering of all binary strings.
Then
$\kpc{x}{y}\equa\kpc{x}{n}\equa 0$,
$\kpm{x}{y}\equa\KP(x)$,
and
we have
$\min_l\{\kpc{x}{y_{1:l}}+\KP(l)\}\equa\KP(x)$
since
$\kpm{x}{y}\leqa\min_l\{\kpc{x}{y_{1:l}}+\KP(l)\}\leqa
\kpc{x}{n}+\KP(n)\equa\KP(x)$.

To get the second example, for an arbitrary $x$
let us take $n$ such that $\KP(l)\ge\KP(x)$ for all $l\ge n$.
Then put $y=0^n1\tilde x$, where $\tilde x$ is any prefix code of $x$
(e.\,g.\,, $\tilde x=0^{\l(x)}1x$).
Obviously, $\kpc{x}{y}\equa 0$ and $\kpm{x}{y}\equa 0$.
Consider $\kpc{x}{y_{1:l}}+\KP(l)$.
If $l\le n$, then it is equal to
$\kpc{x}{0^l}+\KP(l)\geqa\KP(\pair{x,l})\geqa\KP(x)$.
If $l>n$, then $\KP(l)\ge\KP(x)$ by definition of $n$.
\qedx\end{proof}

\begin{corollary}
The future deviation of $M_t$ from $\mu_t$ is bounded by
\bqan
  \sum_{t=l+1}^\infty \E[s_t|\o_{1:l}]
  & \leqa &
  [\kpm{\mu}{\o_{1:l}}+\KP(\lceil d_\mu(\o_{1:l})\rceil)]\ln 2
\\
  & \leqa &
  [\min_{i\le l}\{\kpc{\mu}{\o_{1:i}}\!+\!\KP(i)\}
   +\KP(\lceil d_\mu(\o_{1:l})\rceil)]\ln 2\,.
\eqan
\end{corollary}

Let us note that if $\o$ is $\mu$-random, then
${\KP(\lceil d_\mu(\o_{1:l})\rceil)}\leqa
{\KP(\lceil d_\mu(\o_{1:\infty})\rceil)}+{\KP(\KP(\mu))}$,
and therefore we get the bound, which does not increase
with $l$, in contrast to the bound~$(i)$
in~Corollary~\ref{cor:lbnd}.

Finally, let us point out one more approach
to defining the complexity $\KPM$.
The survey~\cite{UspShen:96} provides ``encoding-free'' definitions
of the main complexities.
In a similar fashion, $\KPM$ could be defined as a minimal
(up to an additive constant) function with the following properties:
\begin{enumerate}
\item The function $\kpm{y}{x}$ is non-negative and co-enumerable;
\item $\kpm{y}{xz}\leq\kpm{y}{x}$ for all $x$, $y$, $z$;
\item $\sum\limits_{y}2^{-\kpm{y}{x}}\le 1$ for all $x$.
\end{enumerate}
Probably, condition~2 expressing strict monotonicity is superfluous,
and both conditions~2 and~3 can be replaced by
\begin{itemize}
\item[$2'$.] For any set $A=\{\pair{x,y}\}$ such that
all the first elements $x$ of the pairs from $A$ have a common prolongation
and the second elements $y$ are different for all pairs from $A$,
it holds $\sum\limits_{\pair{x,y}\in A}2^{-\kpm{y}{x}}\le 1$.
\end{itemize}
It is easy to check that these properties are satisfied for all
previously defined ``versions'' of $\KPM$.
We conjecture that all the definitions are equivalent,
though we cannot prove this.

\section{Proof of Theorem~\protect\ref{thm:KPMbound}}\label{secKpmProof}

If $\mu(x)=0$, then $d_\mu(x)=\infty$ and the bound trivially holds.
Below assume that $\mu(x)\ne 0$ and thus $d_\mu(x)$ is finite.

The plan is to get a statement of the form $2^{d}\mu(xy)\leqm M(xy)$,
where $d\approx d_\mu(x)=\lb\frac{M(x)}{\mu(x)}$.
To this end, we define a new semimeasure $\nu$:
we take the set $S=\{z|d_\mu(z)>d\}$
and put $\nu$ to be $2^d\mu$ on prolongations of $z\in  S$;
this is possible since $S$ has $\mu$-measure $2^{-d}$.
Then we have $\nu(z)\le C\cdot M(z)$ by universality of $M$.
However, the constant $C$ depends on $\mu$ and also on $d$.
To make the dependence explicit,
we repeat the above construction for all numbers $d$
and all semimeasures $\mu^T$,
obtaining semimeasures $\nu_{d,T}$,
and take $\nu=\sum 2^{-\KP(d)}\cdot 2^{-\KP(T)}\nu_{d,T}$.
This construction would give us the term $\KP(\mu)$
in the right-hand side of Theorem~\ref{thm:KPMbound}.
To get $\kpm{\mu}{x}$,
we need a more complicated strategy:
instead of a sum of semimeasures $\nu_{d,T}$,
for every fixed $d$
we sum ``pieces'' of $\nu_{d,T}$ at each point $z$,
with coefficients depending on $z$ as well as on $d$ and $T$.

Now proceed with the formal proof.
Let $\{\mu^T\}_{T\in\SetN}$ be any
(effective) enumeration of all enumerable
semimeasures.
For any integer $d$ and any $T$, put
\beqn
   S_{d,T} \;:=\; \{z\mid \sum_{v\in\X^{\l(z)}\setminus\{z\}}
                 \mu^T(v) + 2^{-d}M(z) > 1\}\,.
\eeqn
The set $S_{d,T}$ is enumerable given $d$ and $T$.

Let $E$ be the optimal $\KPM$-correct set
(satisfying all three requirements),
$E(p,z)$ is the corresponding partial computable function.
For any $z\in\X^*$ and $T$, put
\beqn
  \lambda_{d,T}(z) \;:=\;
  \max\{2^{-\l(p)}\mid\exists k\le\l(z)\colon
   z_{1:k}\in S_{d,T}\text{ and } E(p,z_{1:k})=T\}
\eeqn
(if there is no such $p$, then $\lambda_{d,T}(z)=0$).
Put
\beqn
  \tilde\nu_d(z) \;:=\; \sum_{T}
  \lambda_{d,T}(z)\cdot 2^{d}\mu^{T}(z)\,.
\eeqn
Obviously, this value is enumerable.
It is not a semimeasure, but it has the following property.%
\begin{claim}\label{claim:prefix}
For any prefix-free set $A$,
\beqn
  \sum_{z\in A}\tilde\nu_d(z) \;\le\; 1\,.
\eeqn
\end{claim}
This implies that there exists an enumerable semimeasure $\nu_d$
such that $\nu_d(z)\ge\tilde\nu_d(z)$ for all $z$.
Actually, to enumerate $\nu_d$,
one enumerates $\tilde\nu_d(z)$ for all $z$,
and at each step the current approximation of $\nu_d(z)$
is the maximum of the current approximations of $\tilde\nu_d(z)$
and $\sum_{u\in\X}\nu_d(zu)$.
Trivially, this provides $\nu_d(z)\ge\sum_{u\in\X}\nu_d(zu)$.
To show that $\nu_d(\emptyword)\le 1$,
let us note that at any step of enumeration the current approximation
of $\nu_d(\emptyword)$
is the sum of current approximations of $\tilde\nu_d(z)$ over some
prefix-free set, and thus is bounded by $1$.
Put
\beqn
  \nu(z) \;:=\; \sum_{d} 2^{-\KP(d)}\nu_{d}(z)\,.
\eeqn
Clearly, $\nu$ is an enumerable semimeasure,
thus $\nu(z)\leqm M(z)$.
Let $\mu$ be an arbitrary computable measure,
and
$x,y\in\X^*$.
Let $p\in\fin$ be a string such that $\kpm{\mu}{x}=\l(p)$,
$E(p,x)=T$, and $\mu=\mu^T$.
Put $d=\lceil d_\mu(x)\rceil-1$, i.e.,
${d_\mu(x)-1}\le d < d_\mu(x)$. Hence $\mu(x) < 2^{-d}M(x)$.
Since $\mu=\mu^{T}$ is a measure,
we have
$\sum_{v\in\X^{\l(x)}} \mu^T(v)=1$,
and therefore $x\in S_{d,T}$.
By definition, $\lambda_{d,T}(xy)\ge 2^{-\l(p)}$,
thus $\tilde\nu_{d}(xy)\ge 2^{-\l(p)}2^d\mu(xy)$,
and
\beqn
  2^{-\KP(d)} 2^{-\l(p)} 2^d\mu(xy)
 \;\le\; \nu(xy) \;\leqm\; M(xy)\,.
\eeqn
Replacing $2^d$ in the left-hand side by a smaller value
$2^{d_\mu(x)-1}$,
after trivial transformations we get
\beqn
  \lb\frac{\mu(y|x)}{M(y|x)} \;\leqa \kpm{\mu}{x}+\KP(d)\,,
\eeqn
which completes the proof of Theorem~\ref{thm:KPMbound}.

\renewcommand{\proofname}{\bfseries Proof of Claim~\ref{claim:prefix}}
\begin{proof}
First observe that for all $z\in S_{d,T}$
\beqn
 M(z) > 2^d \mu^T(z)\,,
\eeqn
since
\beqn
\sum_{v\in\X^{\l(z)}\setminus\{z\}}
   \mu^T (v)
 + 2^{-d}M(z) > 1
\quad\text{and}\quad
\sum_{v\in\X^{\l(z)}} \mu^T (v) \le 1
\eeqn
by definition of $S_{d,T}$ and by the semimeasure property,
respectively.
To prove the claim we will group items with the same $\mu^T$,
replace sums of $\mu^T$-measures of several $z$ by the $\mu^T$-measure
of their common prefix from $S_{d,T}$, change $\mu^T$ to $M$
using the inequality above,
and finally show (using ``prefix-free'' properties of $\KPM$)
that the coefficients of $M(z)$ in the sum are small.
By definition,
\beqn
\sum_{z\in A}\tilde\nu_d(z)=
\sum_{z\in A}\sum_{T}
  \lambda_{d,T}(z)\cdot 2^{d}\mu^T(z)=
\sum_{T}\sum_{z\in A}
  \lambda_{d,T}(z)\cdot 2^{d}\mu^T(z)\,.
\eeqn
Let us estimate the inner sum.
Let $\pi_{d,T}(z)$ be the string $p$ that gives the maximum
in the definition of $\lambda_{d,T}(z)$
(if there are several such $p$ we always take,
say, the lexicographically first),
that is $\lambda_{d,T}(z)=2^{-\l(p)}$
and there exists $z'$ being a prefix of $z$
such that $z'\in S_{d,T}$ and $E(p,z')=T$.
Let $\zeta_{d,T}(z)$ be the shortest of such $z'$.
It is easy to see that
$\zeta_{d,T}(\zeta_{d,T}(z))=\zeta_{d,T}(z)$
and $\lambda_{d,T}(\zeta_{d,T}(z))=\lambda_{d,T}(z)$.
\bqan
\sum_{z\in A} \lambda_{d,T}(z)\cdot 2^{d}\mu^T(z)
&=&
\sum_{v}\sum_{z\in A:\zeta_{d,T}(z)=v\hspace{-6ex}}
  \lambda_{d,T}(z)\cdot 2^{d}\mu^T(z)
\;=\; \sum_{v}\sum_{z\in A:\zeta_{d,T}(z)=v\hspace{-6ex}}
  \lambda_{d,T}(v)\cdot 2^{d}\mu^T(z)
\\
 &\leq& \sum_{v\colon \exists z\in A:\zeta_{d,T}(z)=v\hspace{-8ex}}
  \lambda_{d,T}(v)\cdot 2^{d}\mu^T(v)
\;\leq\; \sum_{v\colon \zeta_{d,T}(v)=v\hspace{-3ex}}
  \lambda_{d,T}(v)\cdot 2^{d}\mu^T(v)
\\ &<& \sum_{v\colon \zeta_{d,T}(v)=v\hspace{-3ex}}
  \lambda_{d,T}(v) M(v)\,.
\eqan
In the first inequality we used that
$\zeta_{d,T}(z)$ is a prefix of $z$,
that the set $A$ is prefix free, and
summed the $\mu^T(z)$ to $\mu^T(v)$.
Now we can forget about $A$.
If $\zeta_{d,T}(z)=v$ for some $z$, then
$\zeta_{d,T}(v)=\zeta_{d,T}(\zeta_{d,T}(z))=v$,
and we get the second inequality.
The last inequality holds since $\zeta_{d,T}(v)$
belongs to $S_{d,T}$.
Thus, we need to bound the sum
\beqn
  \sum_T\sum_{v\colon v=\zeta_{d,T}(v)} \lambda_{d,T}(v) M(v)
  \;=\; \sum_v\left(\sum_{T\colon v=\zeta_{d,T}(v)\nq\nq} \lambda_{d,T}(v)\right) M(v)\,.
\eeqn

We say that a function $f\colon\X^*\to[0,1]$
is \emph{unit-summable along any sequence}
if for any $z\in\X^*$
\beqn
  \sum_{i=1}^{\l(z)} f(z_{1:i}) \;\le\; 1\,.
\eeqn

\begin{claim}\label{claim:conv}
The function $f(v)=\sum\limits_{T\colon v=\zeta_{d,T}(v)\hspace{-1em}}\lambda_{d,T}(v)$
is unit-summable along any sequence.
\end{claim}

\begin{lemma}\label{lem:sum}
Let $\nu$ be a semimeasure.
If a function $f$ is unit-summable along any sequence,
then
\beqn
  \sum_{z\in\X^*} f(z)\nu(z) \;\le\; 1\,.
\eeqn
\end{lemma}
This concludes the proof of Claim~\ref{claim:prefix}.
\qedx\end{proof}
\renewcommand{\proofname}{\standardproofname}

\renewcommand{\proofname}{\bfseries Proof of Lemma~\ref{lem:sum}}
\begin{proof}
Since $f(z)$ and $\nu(z)$ are non-negative,
it is sufficient to prove
$\sum_{\l(z)\le n}f(z)\nu(z)\le 1$ for all $n$.
Also we can assume that $\nu$ is a measure
(the sum does not decrease, if $\nu$ is increased to a measure).
\bqan
  \sum_{\l(z)\le n}f(z)\nu(z)
  &\;=\;& \sum_{\l(z)\le n}f(z)
 \sum_{\substack{\l(v)=n,\\ \text{$z$ prefix of $v$}}}\nu(v)
  \;=\; \sum_{\l(v) = n}
 \sum_{\substack{\l(z)\le n,\\ \text{$z$ prefix of $v$}}}f(z)\nu(v)
\\
 &\;=\;& \sum_{\l(v)=n}\sum_{i=1}^n f(v_{1:i})\nu(v)
 \;\le\; \sum_{\l(v)=n}\nu(v)\le 1\,.
\eqan
\qedx\end{proof}
\renewcommand{\proofname}{\standardproofname}

\renewcommand{\proofname}{\bfseries Proof of Claim~\ref{claim:conv}}
\begin{proof}
Take any $z\in\X^*$. Let us show that
\beqn
  \sum_{\substack{ v\text{ prefix of }z,\\
                 T\colon v=\zeta_{d,T}(v)}} \lambda_{d,T}(v)
  \;\le\; 1\,.
\eeqn
Recall that if $\lambda_{d,T}(v)\ne 0$, then
$\lambda_{d,T}(v)=2^{-\l(\pi_{d,T}(v))}$.
We will show that the set
$B(z)=\{\pi_{d,T}(v)\mid v=\zeta_{d,T}(v), \text{ $v$ is a prefix of }z\}$
is prefix free,
and if
$\pi_{d,T_1}(v_1)=\pi_{d,T_2}(v_2)\in B(z)$,
then $v_1=v_2$ and $T_1=T_2$.
Consequently,
\beqn
  \sum_{\substack{ v\text{ prefix of }z,\\
                   T\colon v=\zeta_{d,T}(v)}} \lambda_{d,T}(v)
  \;=\; \sum_{p\in B(z)} 2^{-\l(p)} \;\le\; 1\,.
\eeqn
Assume the converse, that there exist
different $v_i$, $T_i$, $i=1,2$,
such that
$p_1=\pi_{d,T_1}(v_1)$ is a prefix
(proper or not) of $p_2=\pi_{d,T_2}(v_2)$,
$v_1$ and $v_2$ are prefixes of $z$,
and $v_i=\zeta_{d,T_i}(v_i)$.

By definition of $\zeta$,
we have $v_i\in S_{d,T_i}$
and $T_i=E(p_i,v_i)$.
Hence, by the second requirement of $\KPM$-correctness,
$T_1=E(p_1,v_1)=E(p_2,z)=E(p_2,v_2)=T_2$.
Let $T=T_1=T_2$.

Let us show that $v_1=v_2$ too.
Since they both are prefixes of $z$,
one of them is a prefix of the other.
Suppose $v_1$ is a prefix of $v_2$:
By the second requirement of $\KPM$-correctness,
$E(p_2,v_1)=E(p_1,v_1)=T$.
By definition, $\zeta_{d,T}(v_2)$ is the shortest prefix
of $v_2$ belonging to $S_{d,T}$ and such that $E(p_2,\cdot)=T$,
therefore $\zeta_{d,T}(v_2)$ is a prefix of $v_1$,
and thus $v_1=v_2$.
Suppose $v_2$ is a prefix of $v_1$.
Since $E(p_1,v_1)=T$ and $E(p_2,v_2)=T$,
we have $E(p_1,v_2)=T$ by the third requirement of $\KPM$-correctness.
As before, we get $\zeta_{d,T}(v_1)$ is a prefix of $v_2$,
and $v_1=v_2$.
\qedx\end{proof}
\renewcommand{\proofname}{\standardproofname}

\section{Discussion}\label{secDisc}

\paradot{Conclusion}
We evaluated the quality of predicting a stochastic sequence at an
intermediate time, when some beginning of the sequence has been
already observed, estimating the future loss of the universal
Solomonoff predictor $M$. We proved general upper bounds for the
discrepancy between conditional values of the predictor $M$ and
the true environment $\mu$, and demonstrated a kind of tightness
for these bounds. One of the bounds is based on a new variant of
conditional algorithmic complexity $\KPM$, which has interesting
properties on its own. In contrast to standard prefix complexity
$\KP$, $\KPM$ is a monotone function of conditions:
$\kpm{y}{xz}\leq\kpm{y}{x}$.

\paradot{General Bayesian posterior bounds}
A natural question is whether posterior bounds for general
Bayes mixtures based on general $\M\ni\mu$ could also be derived.
The mixture representation~(\ref{xidefsp}) can be written
as a posterior representation
\beqn
  \xi(y|x) \;=\; \sum_{\nu\in\M} w_\nu(x)\nu(y|x)
  \;\geq\; w_\mu(x)\mu(y|x), \qmbox{where}
  w_\nu(x) := w_\nu{\nu(x)\over\xi(x)}
\eeqn
is the posterior belief in $\nu$ after observing $x$ (and $w_\nu$ is the prior).
This immediately implies the bound $D_{l:\infty}\leq\ln
w_\mu(\o_{<l})^{-1}$. Strangely enough, for $\M=\M_U$, $\lb
w_\nu^{-1}:=\KP(\nu)$ does \emph{not} imply $\lb
w_\mu(x)^{-1}=\kpc{\mu}{x}$, not even within logarithmic accuracy,
so it was essential to consider $D_{l:\infty}$. It would be
interesting to derive bounds on $D_{l:\infty}$ or $\ln
w_\mu(x)^{-1}$ for general $\M$ similar to the ones derived here
for $\M=\M_U$.

\paradot{Online classification}
All considered distributions $\rho(x)$ (in particular $\xi$, $M$,
and $\mu$) may be replaced everywhere by distributions
$\rho(x|z)$ additionally conditioned on some $z$. The
$z$-conditions nowhere cause problems as they can essentially be
thought of as fixed (or as oracles or spectators).
An (i.i.d.) classification problem is a typical example: At time
$t$ one arranges an experiment $z_t$ (or observes data $z_t$),
then tries to make a prediction, and finally observes the true
outcome $x_t$ with probability $\mu(x_t|z_t)$. In this case
$\M=\{\nu(x_{1:n}|z_{1:n})=\nu(x_1|z_1)\cdot...\cdot\nu(x_n|z_n)\}$.
(Note that $\xi$ is not i.i.d).
Solomonoff's bound $\KP(\mu)\ln 2$ in~(\ref{prBnd}) holds unchanged.
Compared to the sequence prediction case we have extra information
$z$, so we may wonder whether some improved bound $\kpc{\mu}{z}$
or so, holds. For a fixed $z$ this can be achieved by also
replacing $2^{-\KP(\mu)}$ in~(\ref{xidefsp}) by
$2^{-\kpc{\mu}{z}}$. But if at time $t$ only $z_{1:t}$ is known
like in the classification example, this leads to difficulties
($\xi$ is no longer a (semi)measure, which sometimes can be
corrected~\cite{Poland:04mdl2p}).
Alternatively we could keep definition~(\ref{xidefsp}) but apply
it to the (chronologically correctly ordered) sequence
$z_1x_1z_2x_2...$, condition by~(\ref{defBayes})
to $z_{1:t}$, and try to derive improved bounds.

\paradot{More open problems}
Since $D_{1:\infty}$ is finite, one may expect that the tails
$D_{l:\infty}$ tend to $0$ as $l\to\infty$. However, as
Lemma~\ref{lem:lengthtight} implies, this holds only with
probability 1: for some special $\a$ we have even
$D_{l:\infty}(\a_{<l})\geqa{1\over 3}\KP(l)\toinfty{l}\infty$. It
would be very interesting to find a wide class of $\a$ such that
$D_{l:\infty}(\a_{<l})\to 0$. The natural conjecture is that one
should take $\mu$-random $\a$.
Another (probably, closely related) task is
to study the asymptotic behavior of $\kpm{\mu}{\a_{<l}}$.
It is natural to expect that $\kpm{\mu}{\a_{<l}}$ is bounded
by an absolute constant (independent of $\mu$)
for ``most'' $\a$ and for sufficiently large $l$.
Finally, (dis)proving our conjectured equality of the various
definitions of $\KPM$ we gave, would be interesting and useful.

\paradot{Acknowledgements}
The authors are grateful to Andrej Muchnik, Alexander Shen,
and Nikolai Vereshchagin for discussing the history of
the deficiency conservation theorem
(Theorem~\ref{thm:bounddefect}),
and to anonymous referees for useful comments.


\begin{small}

\end{small}

\end{document}